\documentclass[letterpaper, 10 pt, conference]{ieeeconf}  
\usepackage{graphicx} 
\usepackage{stfloats}
\usepackage{amsmath} 
\usepackage{algorithm,algpseudocode}
\usepackage{wrapfig}
\usepackage{amssymb}
\usepackage{graphicx}
\usepackage{wrapfig}
\usepackage{cite}
\usepackage{booktabs}
\usepackage{pifont}
\usepackage{multirow}
\usepackage{booktabs}  
\usepackage{array}
\usepackage{makecell}
\usepackage{pifont}

\IEEEoverridecommandlockouts                              

\title{\LARGE \bf
PRISM: Projection-based Reward Integration for Scene-Aware Real-to-Sim-to-Real Transfer with Few Demonstrations
}

\author{Haowen~Sun, Han~Wang, Chengzhong~Ma, Shaolong~Zhang, Jiawei~Ye, Xingyu~Chen, Xuguang~Lan$^{*}$
\thanks{The authors are with the National Key Laboratory of Human-Machine Hybrid Augmented Intelligence, Institute of Artificial Intelligence and Robotics, Xi'an Jiaotong University, Xi'an, 710049}
\thanks{$^{*}$Correspondence to: Xuguang Lan \{\texttt{xglan@mail.xjtu.edu.cn}\} }
}

\begin{document}

\maketitle
\thispagestyle{empty}
\pagestyle{empty}

\begin{abstract}

Learning from few demonstrations to develop policies robust to variations in robot initial positions and object poses is a problem of significant practical interest in robotics. Compared to imitation learning, which often struggles to generalize from limited samples, reinforcement learning (RL) can autonomously explore to obtain robust behaviors. Training RL agents through direct interaction with the real world is often impractical and unsafe, while building simulation environments requires extensive manual effort, such as designing scenes and crafting task-specific reward functions. To address these challenges, we propose an integrated real-to-sim-to-real pipeline that constructs simulation environments based on expert demonstrations by identifying scene objects from images and retrieving their corresponding 3D models from existing libraries. We introduce a projection-based reward model for RL policy training that is supervised by a vision-language model (VLM) using human-guided object projection relationships as prompts, with the policy further fine-tuned using expert demonstrations. In general, our work focuses on the construction of simulation environments and RL-based policy training, ultimately enabling the deployment of reliable robotic control policies in real-world scenarios.

\end{abstract}

\section{INTRODUCTION}
\label{sec:Introduction}

Developing robust robotic control policies that generalize across variations in robot initial positions and object poses is a core challenge in embodied AI. While learning directly from real-world demonstrations is intuitive and appealing, it faces limitations in scalability, safety, and data availability. Recent advances such as Vision-Language-Action (VLA) models \cite{kim2024openvla, octo_2023, zhao2025cot} and diffusion-based behavior models \cite{liu2024rdt1bdiffusionfoundationmodel, chi2023diffusion, li2024cogact} demonstrate strong generalization and task understanding, but they demand large-scale, high-quality real-world data, posing significant obstacles for practical deployment. On the other hand, reinforcement Learning (RL) training in simulation enables autonomous policy learning through interaction and has proven effective in complex control tasks. However, the mismatch between simulated training environments and real-world deployment (sim-to-real gap) often leads to poor policy transferability.

To address this challenge, the real-to-sim-to-real paradigm has emerged as a promising direction. Rather than relying solely on manually designed simulators, these approaches \cite{torne2024reconciling, liu2020real, wang2023real2sim2real} leverages real-world data to construct simulation environments that more accurately reflect reality, thereby improving the quality of learned policies and their real-world applicability.
Recent methods employ techniques such as Gaussian Splatting to reconstruct realistic scenes \cite{kerbl20233d, qureshi2024splatsimzeroshotsim2realtransfer, li2024robogsim}, or utilize object configuration files \cite{chen2024urdformer} to build controllable, task-specific environments for RL training. 
While these approaches help reduce domain discrepancies and improve policy stability, they rely on handcrafted reward functions tailored to individual tasks, which hinders generalization and makes scaling to diverse tasks difficult. Furthermore, these methods often underutilize the limited but valuable real-world data, typically incorporating it only as supplemental demonstrations into the RL replay buffer without addressing the fundamental mismatch between simulation and reality.

\begin{figure}
    \centering
    \includegraphics[width=0.5\textwidth]{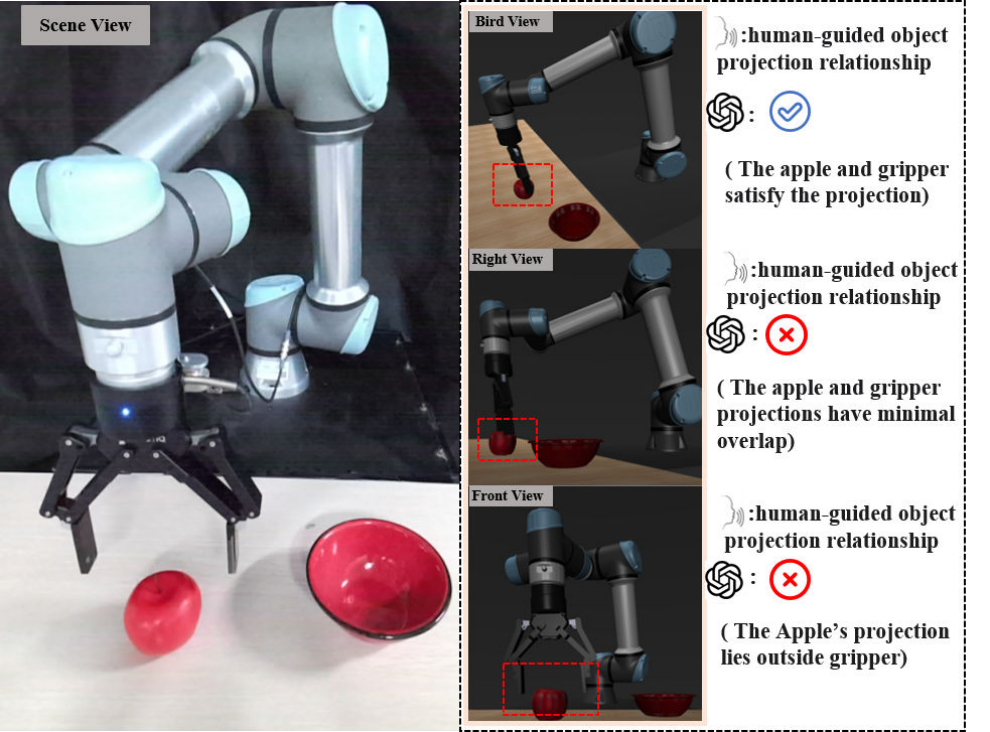}
    \caption{\textbf{Human-guided Object Projection Relationships.} Multi-view simulated images are generated from scene view, and human-guided object projection relationships are used as prompts to query the VLM to evaluate whether each view satisfies the task-specific spatial requirement.
    }
    \label{fig0}
    \vspace{-20pt}
\end{figure}

Therefore, we propose PRISM, an approach for robust policy learning from few demonstrations by constructing a simulation environment based on scene images and a projection-based reward model to minimize the gap between real and simulated environments. Specifically, as shown in Figure \ref{fig0}, we utilize human-provided projection relationships, defined as the viewpoint-dependent ordering of the 2D projections of objects caused by mutual occlusion from a human observer’s perspective. By constructing a simulation environment from real-world scene images, we jointly leverage the projection-based reward model and replayed demonstrations for RL policy training. The learned reward model are converted into an action feasibility predictor, and both the predictor and policy are transferred back to real-world scenarios for effective execution.

In summary, we make the following contributions:

\begin{itemize}
    \item We devise a real-to-sim-to-real pipeline that constructs visually and geometrically consistent simulation environments using 3D model libraries, and transfers policies to the real world via co-training with expert demonstrations and an action feasibility predictor.

    \item We introduce a projection-based reward model that transfers human-guided object projection relationships from vision-language foundation models (VLMs) to RL, enabling efficient reward supervision.
    
   \item We demonstrate that PRISM learns robust policies across six manipulation tasks with varying robot initial poses and object configurations, achieving a 68\% higher average success rate than the baseline.
\end{itemize}

\section{Related Work}
\label{sec:Related Work}
\subsection{Policy Learning via Real-to-Sim-to-Real Transfer}
Recent trends in learning visuomotor control from demonstrations have primarily focused on training policies in simulation and transferring them to the real world via sim-to-real techniques\cite{chen2022system, chen2023visual, andrychowicz2020learning, handa2023dextreme}. To reduce the reliance on large amounts of expert demonstrations in the target domain, real-to-sim-to-real approaches have gained increasing attention, leveraging real-world priors to guide simulation and improve transfer efficiency. However, existing real-to-sim-to-real approaches\cite{liu2020real, kerbl20233d, torne2024reconciling, wang2023real2sim2real, lim2022real2sim2real, wu2024rl, ho2021retinagan, chen2023genaug, mandi2022cacti} primarily focus on reconstructing high-fidelity simulation environments from real-world scenes, while paying less attention to how RL should be conducted in simulation to effectively bridge the sim-to-real gap. PRISM leverages scene images and a 3D model library to efficiently reconstruct simulation environments, enabling rapid environment generation. It further makes full use of few expert demonstrations to support stable policy learning.

\subsection{Large Pre-trained Models as Reward Models}
Recent works in reward model learning have explored leveraging pretrained large language models (LLMs) to generate rewards or synthesize structured training code\cite{yu2023language, xie2024text2reward, ma2023eureka, klissarov2023motif} for RL agents. However, LLMs lack the ability to perceive and reason about visual and spatial structures, recent works have turned to VLMs\cite{ma2023liv, ma2024dreureka, rocamonde2023vision, nam2023lift, bai2022constitutional}, which combine semantic understanding with visual perception, enabling them to serve as reward estimators in robotic tasks that require spatial reasoning. Existing methods often rely on multiple rollouts to infer rewards by comparing observations, which may misrepresent the true task reward and introduce noisy supervision. This can destabilize policy training, especially in long-horizon or high-variance settings. PRISM queries a VLM using human-guided object projection relationships as prompts to supervise the learning of a projection-based reward model. To mitigate the impact of potential misjudgments from the VLM, PRISM further leverages multi-view observations available in simulation to verify the consistency of the inferred rewards across different viewpoints.

\begin{figure*}[ht]
    \vspace{2pt}
    \centering
    \includegraphics[width=1\textwidth]{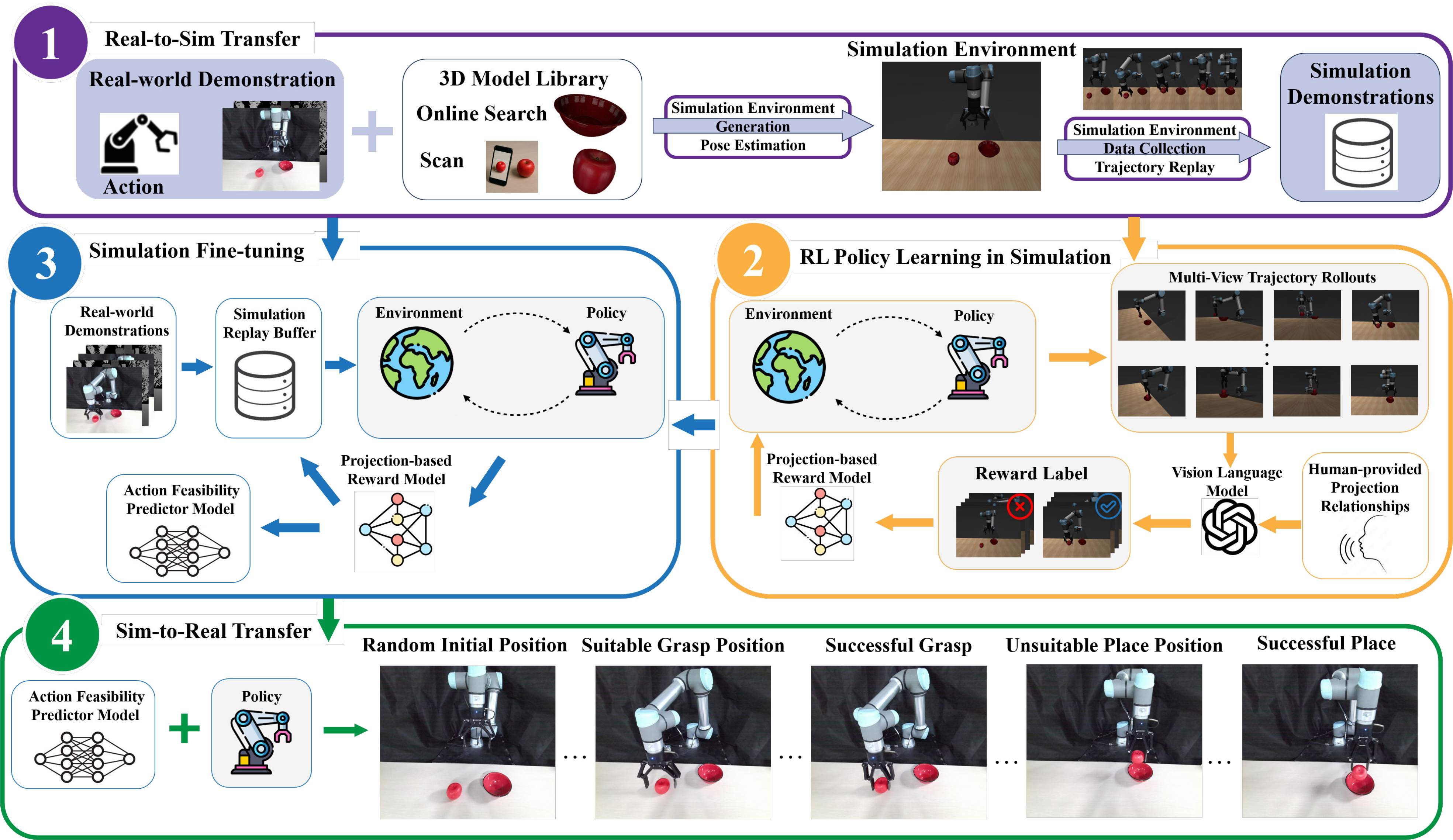}
    \caption{\textbf{PRISM System Overview.} 1) Transfer the real-world scene to the simulator by estimating object poses in the environment and collecting simulation data (see Section \ref{sec:Real-to-Sim Transfer for Scene-Aware Simulation Environment Generation}). 2) Train the reward model using human-guided object projection relationships and apply it to RL. Injecting initialization noise in the simulation enhances the robustness of the control policy (see Section \ref{sec:Robustifying Real-World Imitation Learning Policies in Simulation}). 3) Fine-tune the learned policy from simulation using few real-world demonstrations and train an action feasibility predictor model (see Section \ref{sec: Co-Training on Real-World Data for Sim-to-Real Transfer}). 4) Evaluate policy stability in real-world tasks, ensuring robust behaviors that generalize to novel robot initial states and object poses.}
    \label{fig1}
    \vspace{-12pt}
\end{figure*}

\section{Background}
\label{sec:Background}

\textbf{Simulation environment generation and data collection.} To formalize the problem of simulation scene generation and the transformation of expert demonstrations from  real world to simulation, we consider a 3D model library \( \mathcal{M} = \{M_1, M_2, \dots, M_{N}\} \) and real-world expert demonstration dataset \(
    \mathcal{D}^{\mathrm{real}} = \{\tau_1, \tau_2, \dots, \tau_{K}\},
\)
where each demonstration $\tau_i$ is a sequence of RGB-D observation-action pairs:
\(
    \tau_i = \bigl\{(s_t, a_t)\bigr\}_{t=1}^{H_i},
\)
with $s_t$ representing the observed state at time $t$ and $a_t$ representing the corresponding expert action, and $H_i$ representing the length of trajectory $\tau_i$. Each object $M_m$ in the real-world scene is mapped to the simulation environment via a 3D transform $T \in \mathrm{SE}(3)$.
The simulation environment is represented as \( \mathcal{W} = \bigl\{(M, T) \mid M \in \mathcal{M}, T \in \mathrm{SE}(3) \bigr\} \). Based on the simulation environment \( \mathcal{W} \), we map actions from the real-world expert demonstrations \( \mathcal{D}^{\mathrm{real}} \) into the simulation, producing a simulated dataset \( \mathcal{D}^{\mathrm{sim}} \).

\begin{equation}
\mathcal{D}^{\text{sim}} = \left\{ \hat{\tau}_i \mid \hat{\tau}_i = \left\{ (\hat{s}_t, a_t) \right\}_{t=1}^{H_i}, \;  (s_t, a_t)_{t=1}^{H_i} \in \tau_i \right\}.
\end{equation}
where \( \tau_i \in \mathcal{D}^{\mathrm{real}} \) and \( \hat{s}_t \) represents the simulated state at time \( t \) corresponding to the real-world state \( s_t \).

\textbf{Projection-based reward model.} 
We define human-guided object projection relationships as viewpoint-dependent spatial constraints that emerge when multiple 3D objects are projected onto a 2D image plane, capturing their relative ordering through projection overlap and mutual occlusion from a human observer’s perspective.
By leveraging human-guided object projection relationships as prompts, VLMs generate reward labels conditioned on the current task and a set of \(N_c\) multi-view visual observations \(\{o^j\}_{j=1}^{N_c}\).
Each label \(y \in \{0, 1\}\), where \(y = 1\) indicates that the projection relationships are satisfied across all views, and \(y = 0\) indicates that at least one view violates the relationship.
These VLM-supervised labels are then used to train a projection-based reward model \(r_\psi(\{o^j\}_{j=1}^{N_c}, a)\). Given a dataset of reward labels \( \mathcal{D}^{\text{reward}} = \{(o_{1:N_c}, y)\} \), our algorithm optimizes the reward model \( r_{\psi} \) by minimizing the Binary Cross-Entropy loss.


\textbf{RL Policy and Reward Model Optimization.} Given the simulated demonstration dataset \( \mathcal{D}^{\mathrm{sim}} \), we consider the standard Markov Decision Process (MDP) and RL setup. At each timestep \( t \), the agent receives a state \( s_t \) from the environment and selects an action \( a_t \) according to a policy \( \pi(a_t \mid s_t) \). After executing action \(a_t\), the agent transitions to state \(s_{t+1}\) and receives a reward \(r_t\). The goal of the agent is to maximize the expected return, defined as the discounted sum of rewards:
\(
    R = \sum_{k=0}^{\infty} \gamma^k r_k,
\)
where \( \gamma \in (0,1) \) is the discount factor that balances immediate and future rewards. Inspired by preference-based RL algorithms\cite{lee2021pebble, wang2024rl, christiano2017deep}, we adopt an alternating optimization scheme in which the policy \(\pi_\theta\) and the reward model \(r_\psi\) are updated iteratively. The reward \(r_\psi\) is trained using a dataset of reward labels \(\mathcal{D}^{\text{reward}}\) as described above, while the policy \(\pi_\theta\) is optimized with respect to the learned reward model using standard RL techniques.


\section{Method}
\label{sec:Method}
This section introduces PRISM, a real-to-sim-to-real framework that learns robust control policies from few expert demonstrations using simulation-based RL. By leveraging human-guided object projection relationships, we propose a novel projection-based reward model supervised by a vision-language model (VLM), which not only guides policy learning but also helps bridge the sim-to-real gap. While designed for few-shot settings, PRISM can also enhance the robustness of large pretrained models. An overview of PRISM is shown in Figure \ref{fig1}.

\subsection{Real-to-Sim Transfer for Scene-Aware Simulation Environment Generation}
\label{sec:Real-to-Sim Transfer for Scene-Aware Simulation Environment Generation}
 
The first step of PRISM involves constructing simulated scenes that maintain geometric and visual similarity for policy training.
Given a task description and a scene image, a VLM is employed to decompose the task and identify the relevant target objects required for completion. Corresponding 3D models of these objects are then retrieved through online search or scanning, and assembled into 3D model libraries \( \mathcal{M}\) for downstream use. To enable scene-aware simulation environment generation, we reconstruct task-relevant scenes by grounding these models into simulated environments. This process involves: (i) segmenting the environment based on the first observation frame $o_1$ of the demonstration $\tau_i$, and loading the corresponding 3D meshes from 3D model libraries \( \mathcal{M}\), and (ii) specifying the initial poses of all objects in the scene \( \{ \hat{T}_m \}_{m=1}^{M_{\text{sim}}}
\). We leverage the Segment Anything Model (SAM) \cite{Kirillov_2023_ICCV} for object segmentation, and apply off-the-shelf 6D pose estimation techniques to estimate object poses from RGB-D images of the scene. Our pipeline is method-agnostic, supporting various pose estimation algorithms (e.g. FoundationPose\cite{foundationposewen2024}, SAM-6D\cite{lin2023sam}). Each of these algorithms takes scene RGB-D images as input, along with the target object mesh models, and outputs the corresponding 6D object poses.

The next question is how can we refine pose estimation results by incorporating feasible physical manipulation priors derived from environmental data and expert demonstrations to generate simulation demonstrations? Accurately identifying the pose of target objects is challenging, as estimates from pose estimation algorithms often violate fundamental physical constraints. For instance, objects may intersect with the table surface or fail to adhere to gravity constraints. Similarly, pose estimation errors can lead to failures when mapping actions from the real-world expert demonstration dataset \(\mathcal{D}^{\mathrm{real}}\) into the simulation. 


To mitigate these issues, we refine a set of initially estimated object poses to (i) prevent collisions among objects in the simulated environment and (ii) satisfy trajectory-based constraints, ensuring consistency with expert demonstrations (e.g., objects must be correctly positioned within the gripper during demonstrated actions). To achieve this, we impose two categories of constraints on the object poses.

\textbf{Environment constraints} \( \mathcal{C}_{\text{env}} \) ensure physical plausibility of the simulation environment. These include collision-free placement of all objects throughout the trajectory and stable support—i.e., no object should penetrate the ground or tabletop at any time. These constraints are enforced by the simulator, which continuously monitors physical interactions and returns binary feasibility feedback.


\textbf{Trajectory constraints} \( \mathcal{C}_{\text{traj}} \) enforce consistency with key states and actions observed in expert demonstrations. To simplify the optimization process, we define the set of key states to include only two types: \emph{gripper alignment states} and \emph{task final states}. Both are expressed using the same formulation, requiring that at designated timesteps \( t_g \), the target object \( o_m \) lies within an acceptable distance from the robot’s gripper:
\[
\| x_{\text{gripper}}(t_g) - x_{o_m}(t_g) \| \leq \epsilon_g,
\]
where \( \epsilon_g \) denotes a spatial alignment threshold. Here, \( x_{\text{gripper}}(t_g) \) is the gripper position obtained from the expert trajectory at time \( t_g \), and \( x_{o_m}(t_g) \) is extracted from the simulated environment of the object pose \( \hat{T}_m(t_g) \).

For each expert demonstration \( \tau_i \), we refine object poses over time while ensuring they satisfy physical manipulation priors. To support accurate simulation of real-world behavior, each demonstration is mapped to a dedicated simulation environment for 100\% replay accuracy, producing high-fidelity data \( \mathcal{D}^{\mathrm{sim}} \) for RL training. The refinement is formulated as the following optimization problem:


\begin{equation}
\begin{aligned}
\min_{\{\hat{T}_m^{(i)}(t)\}_{m=1,t=1}^{M_{\text{sim}},T}} \quad 
& \sum_{t=1}^{T} \sum_{m=1}^{M_{\text{sim}}} \| x_{\text{gripper}}^{(i)}(t) - x_{o_m}^{(i)}(t) \|^2 \\
\text{s.t.} \quad 
& \{ \hat{T}_1^{(i)}, \dots, \hat{T}_{M_{\text{sim}}}^{(i)} \} \in \mathcal{C}_{\text{env}}^{(i)} \cap \mathcal{C}_{\text{traj}}^{(i)},
\end{aligned}
\end{equation}



\begin{figure*}[ht]
    \centering
    \includegraphics[width=1\textwidth]{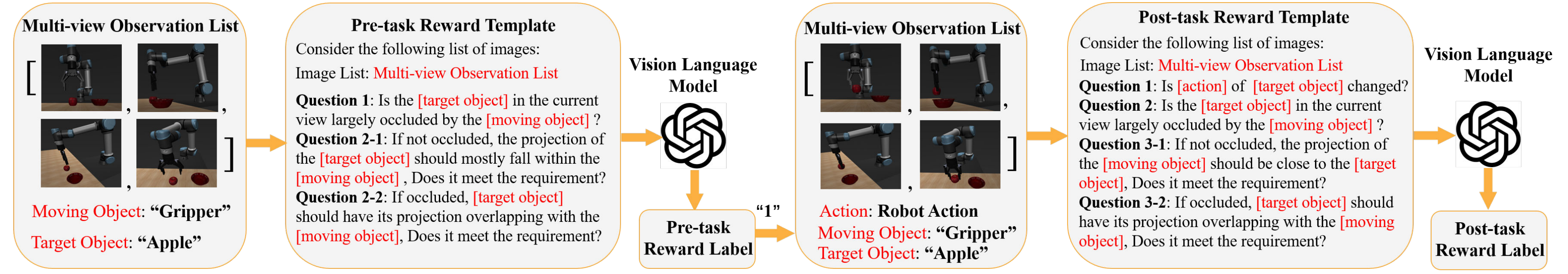}
    \caption{\textbf{Two-stage VLM querying process for task-specific reward labeling.} The pre-task prompt evaluates whether the current state satisfies the condition for executing the action, while the post-task prompt evaluates task completion. This template generalizes across all similar tasks.
    }
    \label{fig2}
    \vspace{-12pt}
\end{figure*}

\subsection{Learning Robust RL Policy in Simulation}
\label{sec:Robustifying Real-World Imitation Learning Policies in Simulation}

The next step in PRISM is to learn a robust policy that can solve desired tasks across a wide range of configurations and environmental conditions. To make efficient use of few expert demonstrations, we employ BC-SAC as the underlying RL algorithm. Traditionally, RL training requires carefully handcrafted reward functions, often demanding prohibitively extensive manual effort for each task. Instead, we adopt an innovative approach by leveraging human-guided object projection relationships, reducing the reliance on manual reward engineering.

We assume that the VLMs have been pretrained on large-scale and diverse image-text corpora, endowing them with broad generalization and reasoning capabilities across a wide range of tasks and environments. We consider large pretrained vision-language foundation models, such as QWen-72B\cite{qwen} and GPT-4 Vision\cite{OpenAI}, to satisfy these assumptions. To mitigate potential errors in VLM judgments, we leverage the simulation environment's ability to preset multiple camera viewpoints, utilizing multi-view observations to improve the accuracy of the assessments.

To train the projection-based reward model and the control policy, we first need to generate reward labels from the VLM. As illustrated in Figure \ref{fig2}, the querying process for each task consists of two stages: pre-task reward querying and post-task reward querying. Pre-task reward querying determines whether the current state meets the conditions required to execute the task, while post-task reward querying evaluates whether the task has been successfully completed. At each stage, the VLM evaluates the corresponding human-provided projection relationships to assign a reward label \( y \in \{0,1\} \), where 1 indicates that all views satisfy the projection relationship and a reward is assigned, while 0 signifies that the projection relationship is violated and no reward is given. The multi-view evaluation iterates over different observations to refine the labels. Finally, we store the reward labels produced by the VLM into the reward label buffer \( \mathcal{D}^{\text{reward}}\) during the training process. Reward learning can then be performed (as detailed in Section \ref{sec:Background}) to supervise the training of the reward model using the buffer \( \mathcal{D}^{\text{reward}}\).


To minimize prompt engineering effort, we adopt a unified template across all skills within the same category (e.g., pick, place, insert). Consequently, training a policy for a new skill requires only specifying the task goal object, while the reward labels and the corresponding reward model are automatically learned through the above process.



\subsection{Co-Training on Real-World Demonstrations for Sim-to-Real Transfer}
\label{sec: Co-Training on Real-World Data for Sim-to-Real Transfer}
During the sim-to-real transfer phase, we incorporate real-world demonstrations into the SAC replay buffer, allowing them to be trained alongside simulated experiences. This approach leverages the complementary strengths of simulation-based robustification and in-domain real-world data, enabling the agent to better adapt to real-world dynamics. However, in real-world scenarios, automatic feedback on task completion is often unavailable. Moreover, executing actions in inappropriate contexts, such as closing the gripper when the object is not correctly positioned, can lead to irreversible task failure (e.g., failed pick attempts). 

To address this issue and further improve execution success rates, we introduce an action feasibility predictor \(r_{\phi}^{\text{real}}(o_t^{\text{RGB-D}}, a_t) \in \{0,1\}\) derived from the pre-trained projection-based reward model. Given the difficulty of acquiring multi-view images in real-world environments, we replace multi-camera inputs with single-view RGB-D observations. The depth channel provides geometric cues that allow the model to infer 3D spatial relationships, effectively approximating the multi-view reward model used during simulation.
\subsection{ Implementation Details}
\label{sec: Implementation Details}
For policy training, we adopt a BC-SAC framework that takes single-view RGB observations as input. Following PEBBLE~\cite{lee2021pebble}, we relabel all rollouts in the SAC replay buffer once the reward model \( r_\psi \) is updated. Our projection-based reward model \(r_\psi\) takes multi-view RGB images as input and outputs binary reward labels. We use a pretrained DINO vision transformer~\cite{caron2021emerging} as the image encoder, followed by a lightweight MLP head for reward prediction. The action feasibility predictor  \(r_{\phi}^{\text{real}}\) operates on single-view RGB-D observations. For encoding RGB-D, we adopt a U-Net architecture with separate encoders for the RGB and depth channels, followed by feature fusion layers and a binary classification head.

\begin{figure*}[ht]
    \vspace{2pt}
    \centering
    \includegraphics[width=1\textwidth]{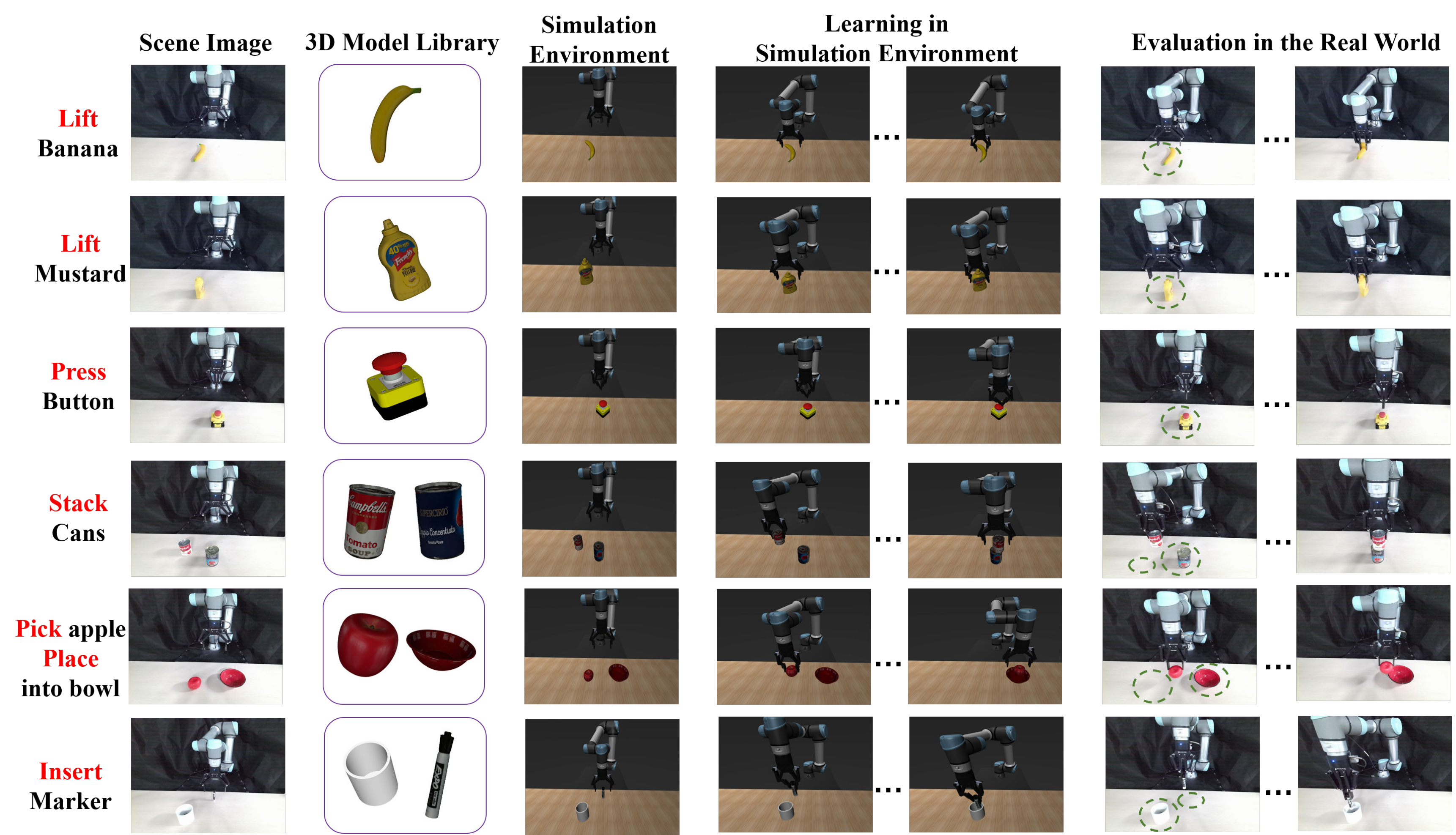}
    \caption{\textbf{Qualitative Results for Real-world Robot Experiments.} The simulation environment is constructed based on RGB-D images and a 3D model library, with additional noise injected to facilitate the learning of a robust control policy. The policy is evaluated on six tasks in the real world. The green dashed lines denote the approximate placement regions for object randomization.
    }
    \label{fig3}
\end{figure*}

\section{Experiments}
\label{sec:result}
Our experiments are designed to answer the following questions about PRISM:
(a) Does PRISM learn robust policies with respect to variations in robot initial states and object poses? 
(b) Is human-guided object projection relationships necessary for policy learning and Sim-to-Real transfer?
(c) What is the accuracy of VLM reward labeling?

To answer these questions, we evaluate PRISM on six different tasks, as shown in Figure \ref{fig3}. These tasks include lifting banana, lifting mustard, pressing button, inserting marker pen, picking apple and placing into bowl, and stacking cans. For each task, we consider randomizing the initial positions of both the object and the robot’s end effector. During each rollout, random noise is added to the end effector’s pose, with object placements perturbed by a translation of \(\pm10 cm\), corresponding to the noise introduced in the simulation environment.

We conduct our experiments using a UR5e robotic arm equipped with a Robotiq-140 gripper, employing 6-DoF Cartesian end-effector position control, and an Azure Kinect camera as the image sensor. In our simulation environment, the robot base coordinate frame serves as the origin. Initial pose estimation is obtained using the FoundationPose \cite{foundationposewen2024} pose estimation algorithm, which is transformed into the world coordinate system based on hand-eye calibration results. The multi-view cameras used for reward model learning include the scene-view, right-view, front-view, and bird-view. The scene-view camera position is also determined through hand-eye calibration, while additional cameras are manually positioned relative to the base coordinate system. All real-world evaluations use the best policy obtained for each method, with results reported from 10 rollouts.


\begin{table*}[t]
    \centering
    \renewcommand{\arraystretch}{1.3}
    \setlength{\tabcolsep}{0.1pt}
    \begin{tabular}{
        m{2.4cm}<{\centering}  
        *{2}{m{0.9cm}<{\centering}}  
        *{2}{m{0.9cm}<{\centering}}  
        *{2}{m{0.9cm}<{\centering}}  
        *{2}{m{0.9cm}<{\centering}}  
        *{2}{m{0.9cm}<{\centering}}  
        *{2}{m{0.9cm}<{\centering}}  
        *{2}{m{0.9cm}<{\centering}}  
    }
        \toprule
        & \multicolumn{2}{c}{\textbf{Lift banana}} 
        & \multicolumn{2}{c}{\textbf{Lift mustard}} 
        & \multicolumn{2}{c}{\textbf{Press button}} 
        & \multicolumn{2}{c}{\textbf{Insert pen}} 
        & \multicolumn{2}{c}{\makecell[c]{\textbf{Pick apple} \\ \textbf{place into bowl}}}
        & \multicolumn{2}{c}{\textbf{Stack cans}} 
        & \multicolumn{2}{c}{\textbf{Average}} \\
        \cmidrule(lr){2-3}
        \cmidrule(lr){4-5}
        \cmidrule(lr){6-7}
        \cmidrule(lr){8-9}
        \cmidrule(lr){10-11}
        \cmidrule(lr){12-13}
        \cmidrule(lr){14-15}
        \textbf{Randomization}
        & \textbf{\ding{55}} & \textbf{\ding{51}}
        & \textbf{\ding{55}} & \textbf{\ding{51}}
        & \textbf{\ding{55}} & \textbf{\ding{51}}
        & \textbf{\ding{55}} & \textbf{\ding{51}}
        & \textbf{\ding{55}} & \textbf{\ding{51}}
        & \textbf{\ding{55}} & \textbf{\ding{51}}
        & \textbf{\ding{55}} & \textbf{\ding{51}} \\
        \midrule
        \textbf{BC (5 demos)} 
        & 3/10 & 2/10 & 2/10 & 1/10 & 4/10 & 3/10 & 3/10 & 2/10 & 0/10 & 0/10 & 0/10 & 0/10 & 12/60 & 8/60 \\

        \textbf{BC (15 demos)} 
        & 8/10 & 4/10 & 6/10 & 3/10 & 5/10 & 3/10 & 4/10 & 4/10 & 3/10 & 1/10 & 2/10 & 3/10 & 28/60 & 18/60 \\

        \textbf{OpenVLA~\cite{kim2024openvla}} 
        & 2/10 & 1/10 & 0/10 & 1/10 & 0/10 & 0/10 & 0/10 & 0/10 & 0/10 & 0/10 & 0/10 & 0/10 & 2/60 & 2/60 \\

        \textbf{OCTO~\cite{octo_2023}} 
        & 4/10 & 5/10 & 6/10 & 4/10 & 4/10 & 4/10 & 2/10 & 3/10 & 1/10 & 1/10 & 0/10 & 1/10 & 17/60 & 18/60 \\

        \textbf{PRISM (5 demos) w/o co-training} 
        & 3/10 & 4/10 & 7/10 & 4/10 & 5/10 & 2/10 & 2/10 & 1/10 & 2/10 & 1/10 & 3/10 & 1/10 & 22/60 & 13/60 \\

        \textbf{PRISM (5 demos)} 
        & \textbf{9/10} & \textbf{10/10} & \textbf{9/10} & \textbf{8/10} & \textbf{6/10} & \textbf{6/10} & \textbf{10/10} & \textbf{9/10} & \textbf{10/10} & \textbf{8/10} & \textbf{9/10} & \textbf{8/10} & \textbf{53/60} & \textbf{49/60} \\

        \bottomrule
    \end{tabular}
    \caption{\textbf{Task success rates under clean and distractor initial conditions.}}
    \label{tab1}
    \vspace{-20pt}
\end{table*}

\subsection{Does PRISM learn robust policies with respect to variations in robot initial states and object poses? }
\label{sec:Does PRISM learn robust policies with respect to variations in robot initial states and object poses? }
In this section, we investigate whether PRISM can learn control policies from few expert demonstrations while maintaining robustness to variations in both the robot's initial positions and object poses. We evaluate our real-to-sim-to-real RL pipeline against two types of baselines: (i) standard behavior cloning (BC), trained exclusively on real-world demonstrations, and (ii) the VLA models, OpenVLA\cite{kim2024openvla} and Octo\cite{octo_2023}. We present the results of the PRISM pipeline, which begins with five demonstrations transferred from the real world to a simulated environment, followed by co-training with real-world demonstrations during fine-tuning.

Table~\ref{tab1} shows that PRISM maintains consistently high performance across various tasks under both randomized and non-randomized conditions, achieving an average success rate of 82\% even when both the robot’s initial position and the object’s pose are perturbed. In contrast, the BC baseline achieves only 8 successful trials under the same randomized setting. These results highlight that PRISM provides a level of robustness that cannot be attained through demonstration-based learning alone. We also compared PRISM with two VLA baselines. While both VLA approaches excel at interpreting visual cues through language-conditioned reasoning, they did not perform as well on our tasks, with OpenVLA\cite{kim2024openvla} achieving only a 3\% success rate, while Octo\cite{octo_2023} performed slightly better, attaining a 30\% success rate across all tasks. These results indicate that while VLA models exhibit a certain degree of scene generalization capability, they still struggle to execute tasks in unseen real-world environments.

We also compare PRISM with behavior cloning using 15 demonstrations to highlight that simply adding more data is insufficient to address the challenges inherent in imitation learning. Although increasing the number of demonstrations leads to a 33\% improvement in success rate, the performance still falls significantly short of that achieved by PRISM. This discrepancy stems from limited state coverage in the demonstrations. As execution errors accumulate during rollout, they introduce novel states that the policy has not encountered during training, making it difficult to achieve high success rates in long-horizon tasks such as picking apple into bowl and stacking cans.

To further validate the effectiveness of our co-training strategy on real-world data for sim-to-real transfer, we conducted an additional ablation study. We compared our method with a variant that directly applies the policy trained in simulation to the real world without co-training. The results show that incorporating co-training leads to a 59\% improvement in task success rates. This demonstrates that our approach not only preserves the generalization ability gained from simulation, but also enables the policy to adapt to real-world visual domains more effectively.

\subsection{Is human-guided object projection relationships necessary for policy learning and Sim-to-Real transfer?}
\label{sec: Is human-provided object projection relationship guidance necessary for policy learning and Sim-to-Real transfer?}
\begin{table}[h]
    \centering
    \renewcommand{\arraystretch}{1.5} 
    \setlength{\tabcolsep}{1pt} 
    \begin{tabular}{m{3cm}<{\centering}m{1cm}<{\centering}m{1.5cm}<{\centering}m{2cm}<{\centering}m{2cm}}
        \toprule
        & \textbf{Lift banana} & \textbf{Insert marker pen} & \textbf{Pick apple place into bowl} \\
        \midrule
        \textbf{PRISM w/ projection} & 10/10 & 9/10 & 8/10\\
        \textbf{PRISM w/o projection} & 5/10 & 1/10 & 0/10 \\
        \textbf{PRISM w/o \(r_{\phi}^{\text{real}}\)} & 8/10 & 8/10 & 5/10 \\
        \textbf{OCTO~\cite{octo_2023} w/ \(r_{\phi}^{\text{real}}\)} & 9/10 & 5/10 & 5/10 \\
        \bottomrule
    \end{tabular}
    \caption{PRISM performance comparison under different conditions.}
    \label{tab2}
    \vspace{-12pt}
\end{table}

\begin{figure}
    \vspace{2pt}
    \centering
    \includegraphics[width=0.49\textwidth]{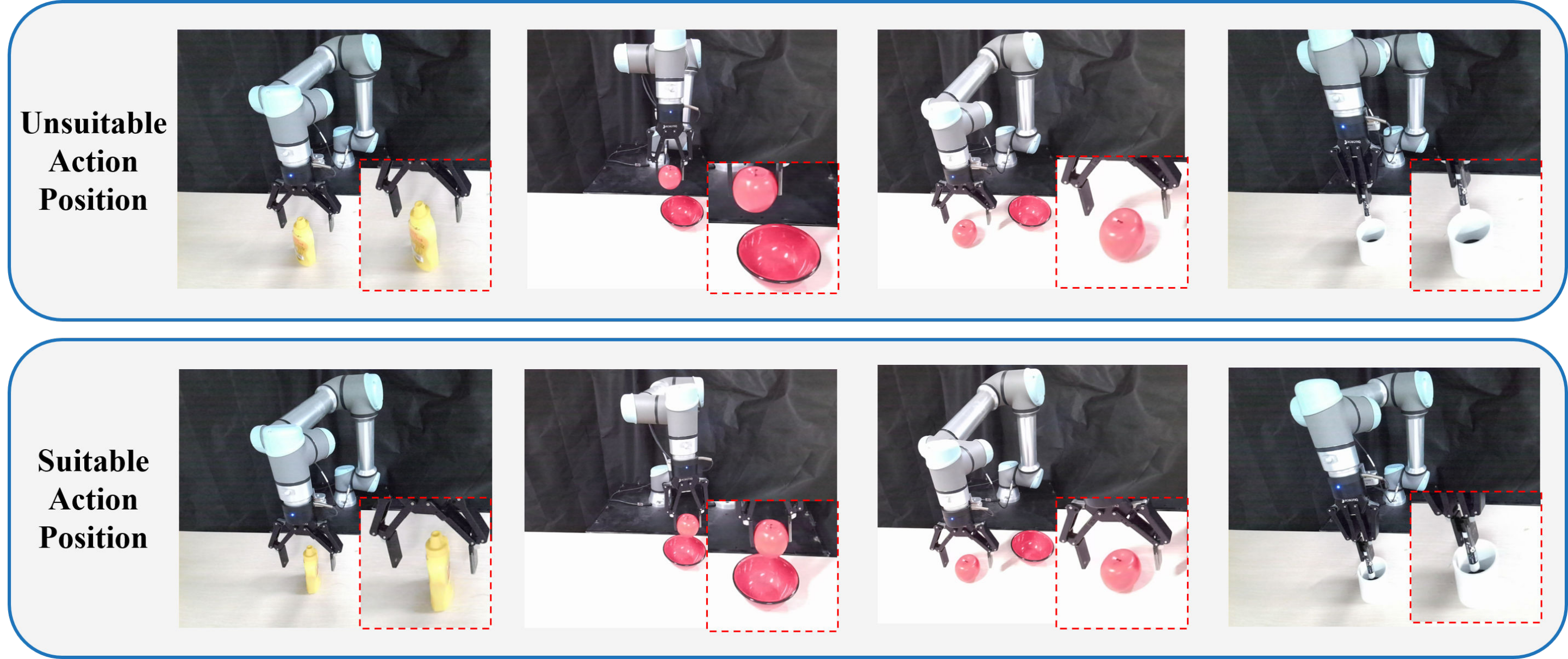}
    \caption{\textbf{Results for Action Feasibility Predictor.} The first row shows positions that typically lead to task failure, while the second row displays positions suitable for successful execution.}
    \label{fig4}
    \vspace{-12pt}
\end{figure}

\textbf{Human-guided object projection relationships in policy learning.}
To evaluate the effectiveness of projection relationships in policy learning, we compare two approaches, as illustrated in Table \ref{tab2}: (i) PRISM guided by human-guided object projection relationships, and (ii) PRISM relying on direct queries to determine the task state. Our results indicate that policies trained with human-guided object projection relationships improve task success rates by nearly 70\%. This performance gain is primarily attributed to the explicit geometric constraints and domain knowledge introduced through the human-provided object projection prompts, which enable the vision-language model to more accurately reason about the spatial configuration of the scene. In contrast, the direct-query approach suffers from limited feedback accuracy. Without human-provided projection relationships as prompts, the VLM lacks sufficient geometric grounding and relies solely on generic priors, which proves inadequate for reliably assessing task states. As a result, this approach yields significantly lower performance and exhibits more unstable training behavior. These experimental findings highlight the critical role of human-guided projection relationships in providing precise and context-aware supervision, ultimately leading to more effective policy learning.

\textbf{Human-guided object projection relationships in sim-to-real Transfer.}
We also investigate the impact of projection relationships in sim-to-real transfer by comparing our action feasibility predictor against directly applying the pipeline to real-world tasks. Specifically, we analyze whether the predictor can prevent failures from inappropriate actions and terminate the task at appropriate states to improve overall success. As shown in Table \ref{tab2}, using the action feasibility predictor improves task execution success rates by 20\%. We further validated this approach by integrating the action feasibility predictor with Octo \cite{octo_2023}, observing a notable 37\% improvement in task success rates in real-world scenarios. Figure \ref{fig4} shows real-world performance when our action feasibility predictor identifies unsuitable positions for executing the next action. These conditions typically correspond to gripper state changes that occur at inappropriate times, such as attempting to grasp when the object is not properly aligned, or failing to terminate the task promptly after successful completion.

\subsection{What is the accuracy of VLM reward labeling?}
\label{sec: What is the accuracy of VLM reward labeling?}

\begin{figure}
    \vspace{2pt}
    \centering
    \includegraphics[width=0.48\textwidth]{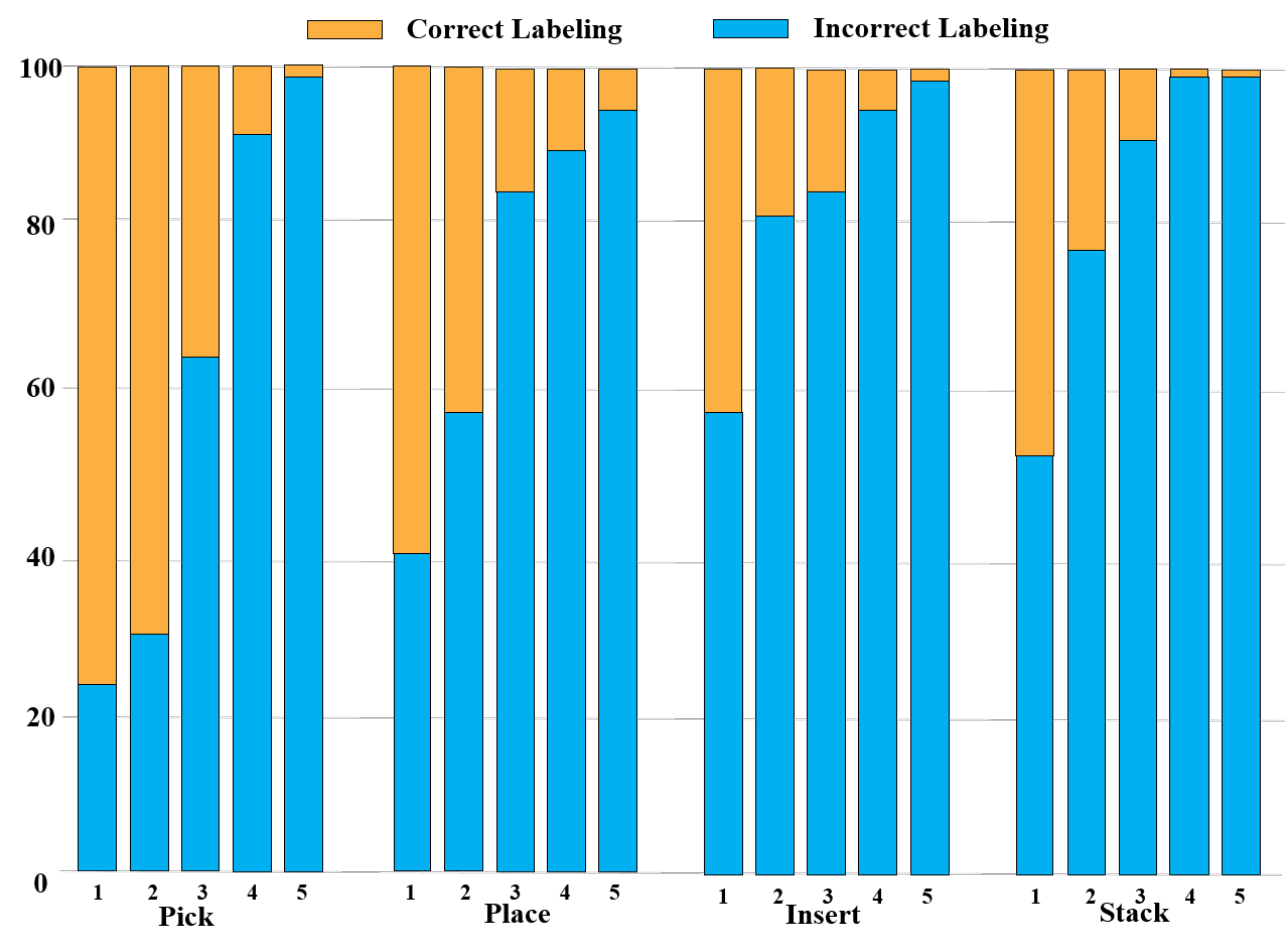}
    \caption{\textbf{The accuracy of VLM reward labels.} The x-axis represents the number of camera viewpoints in the simulation environment, while the y-axis shows the percentage where the VLM preference labels are correct and incorrect. Labeling accuracy improves as the number of views increases.
    }
    \label{fig5}
    \vspace{-2pt}
\end{figure}

Given that PRISM can learn effective rewards and policies to solve tasks, we further analyze the accuracy of the reward labels generated by the VLM. To evaluate accuracy, the VLM produces binary outputs \{\text{0}, \text{1}\}, where "1" indicates that the current state satisfies the condition and is rewarded, while "0" signifies that the condition is not met and no reward is given. We then compare these outputs against ground truth reward labels, which are defined based on the environment's handcrafted reward functions. Our intuition is that, like humans, it would be hard for the VLM to generate accurate reward labels when only a single viewpoint image is available (due to projection ambiguity in occlusion relationships), yet significantly easier to produce correct labels when multi-view images exhibit noticeable dissimilarities in achieving the operational goal. 

We categorize all tasks into four skill types: \textbf{pick}, \textbf{place}, \textbf{insert}, and \textbf{stack}, and evaluate the reward labeling accuracy for each skill type separately. Figure \ref{fig5} demonstrates how the VLM's labeling accuracy evolves with increased viewpoint observations. The x-axis quantifies the number of camera viewpoints from 1 to 5, where each integer represents distinct visual perspectives captured during task execution. The y-axis measures labeling accuracy through dual metrics: blue bars indicate correct reward labels while orange bars denote incorrect judgments. For all skills, we observe a general trend where accuracy improves as the number of viewpoints increases, which aligns with intuition. This trend is particularly clear and consistent across all skills. Overall, across all tasks, we find that the multi-view VLM generates more correct preference labels than incorrect ones, the accuracy of VLM-generated reward labels with four viewpoints is sufficient for learning an effective reward model and policy.

\section{CONCLUSION} 
\label{sec:CONCLUSION}

We propose an integrated end-to-end pipeline PRISM that automatically generates simulation environments from a single scene image and expert demonstrations, leveraging existing 3D model libraries. For RL policy training, we introduce reward models derived from human-guided object projection relationships, with additional guidance from VLMs. The pipeline also supports fine-tuning with expert demonstrations, enabling robust deployment in real-world settings. In general, our work facilitates the construction of automated simulation environments and RL-based policy training, ultimately leading to the deployment of reliable robotic control policies in real-world scenarios.
\addtolength{\textheight}{0cm}   






\bibliographystyle{IEEEtran}
\bibliography{icra}

\end{document}